\documentclass[runningheads]{llncs}

\usepackage{eccv}

\usepackage{eccvabbrv}

\usepackage[dvipsnames]{xcolor}

\usepackage{url}
\usepackage{bm}
\usepackage{algpseudocode}
\usepackage{algorithm}
\usepackage{amsmath}
\usepackage{amssymb}
\usepackage{xcolor}       
\usepackage{xkcdcolors}
\usepackage{colortbl}
\usepackage{wrapfig,lipsum,booktabs}
\usepackage{arydshln}
\usepackage{tabularray}
\usepackage{placeins}
\usepackage{multirow}
\usepackage{booktabs}       
\usepackage{setspace}
\usepackage{siunitx}               
\usepackage[export]{adjustbox}
\usepackage{capt-of}
\usepackage{diagbox}
\usepackage{eso-pic}
\usepackage{epigraph}
\usepackage{pifont}

\makeatletter
\newcommand*\bigdot{\mathpalette\bigdot@{.5}}
\newcommand*\bigdot@[2]{\mathbin{\vcenter{\hbox{\scalebox{#2}{$\m@th#1\bullet$}}}}}
\makeatother

\usepackage{amsmath,amsfonts,bm}

\newcommand{\Ed}{{\mathbb E}}

\def\eqref#1{equation~\ref{#1}}

\def\1{\bm{1}}

\def\rvn{{\mathbf{n}}}

\def\vf{{\bm{f}}}

\def\vm{{\bm{m}}}
\def\vn{{\bm{n}}}

\def\vr{{\bm{r}}}
\def\vs{{\bm{s}}}

\def\vw{{\bm{w}}}
\def\vx{{\bm{x}}}
\def\vy{{\bm{y}}}
\def\vz{{\bm{z}}}

\DeclareMathAlphabet{\mathsfit}{\encodingdefault}{\sfdefault}{m}{sl}
\SetMathAlphabet{\mathsfit}{bold}{\encodingdefault}{\sfdefault}{bx}{n}

\DeclareMathOperator*{\argmin}{arg\,min}

\usepackage{graphicx}
\usepackage{booktabs}
\usepackage{amsmath}
\usepackage{xcolor}       

\usepackage[accsupp]{axessibility}  

\usepackage[pagebackref,breaklinks,colorlinks,citecolor=eccvblue]{hyperref}

\usepackage{orcidlink}

\begin{document}

\title{Prototype Clustered Diffusion Models for Versatile Inverse Problems} 

\titlerunning{Prototype Clustered Diffusion Models for Versatile Inverse Problems}

\author{Jinghao Zhang \and
Zizheng Yang\and
Qi Zhu \and
Feng Zhao 
}

\authorrunning{Jinghao Zhang et al.}

\institute{University of Science and Technology of China \\
\email{\{jhaozhang,yzz6000,zqcrafts\}@mail.ustc.edu.cn} \\
\email{fzhao956@ustc.edu.cn}}

\maketitle

\begin{abstract}
Diffusion models have made remarkable progress in solving various inverse problems, attributing to the generative modeling capability of the data manifold.
Posterior sampling from the conditional score function enable the precious data consistency certified by the measurement-based likelihood term. 
However, most prevailing approaches confined to the deterministic deterioration process of the measurement model, regardless of capricious unpredictable disturbance in real-world sceneries.
To address this obstacle, we show that the measurement-based likelihood can be renovated with restoration-based likelihood via the opposite probabilistic graphic direction, licencing the patronage of various off-the-shelf restoration models and extending the strictly deterministic deterioration process to adaptable clustered processes with the supposed prototype, in what we call restorer guidance.
Particularly, assembled with versatile prototypes optionally, we can resolve inverse problems with bunch of choices for assorted sample quality and realize the proficient deterioration control with assured realistic.
We show that our work can be formally analogous to the transition from classifier guidance to classifier-free guidance in the field of inverse problem solver.
Experiments on multifarious inverse problems demonstrate the effectiveness of our method, including image dehazing, rain streak removal, and motion deblurring. 
\end{abstract}

\section{Introduction}
\label{sec:intro}
Diffusion models~\cite{sohl2015deep,ho2020denoising,song2020score} have recently emerged as impressive generative models with promising performance on various applications such as image generation~\cite{rombach2022high,zhang2023adding,saharia2022photorealistic}, image editing~\cite{meng2021sdedit,brooks2023instructpix2pix,ruiz2023dreambooth}, video generation~\cite{ho2022imagen}, speech synthesis~\cite{huang2022fastdiff}, and 3D generative modeling~\cite{poole2022dreamfusion,tewari2024diffusion}.
Apart from that, diffusion models are also served as competitive candidates for inverse problem solver, which aim at reversing the deterioration process from the contaminated measurement $\vy$ to original complete signal $\vx$~\cite{chung2022improving,chung2022diffusion,song2022pseudoinverse}.

\begin{figure*}
  \begin{center}
  \includegraphics[width=0.76\textwidth]{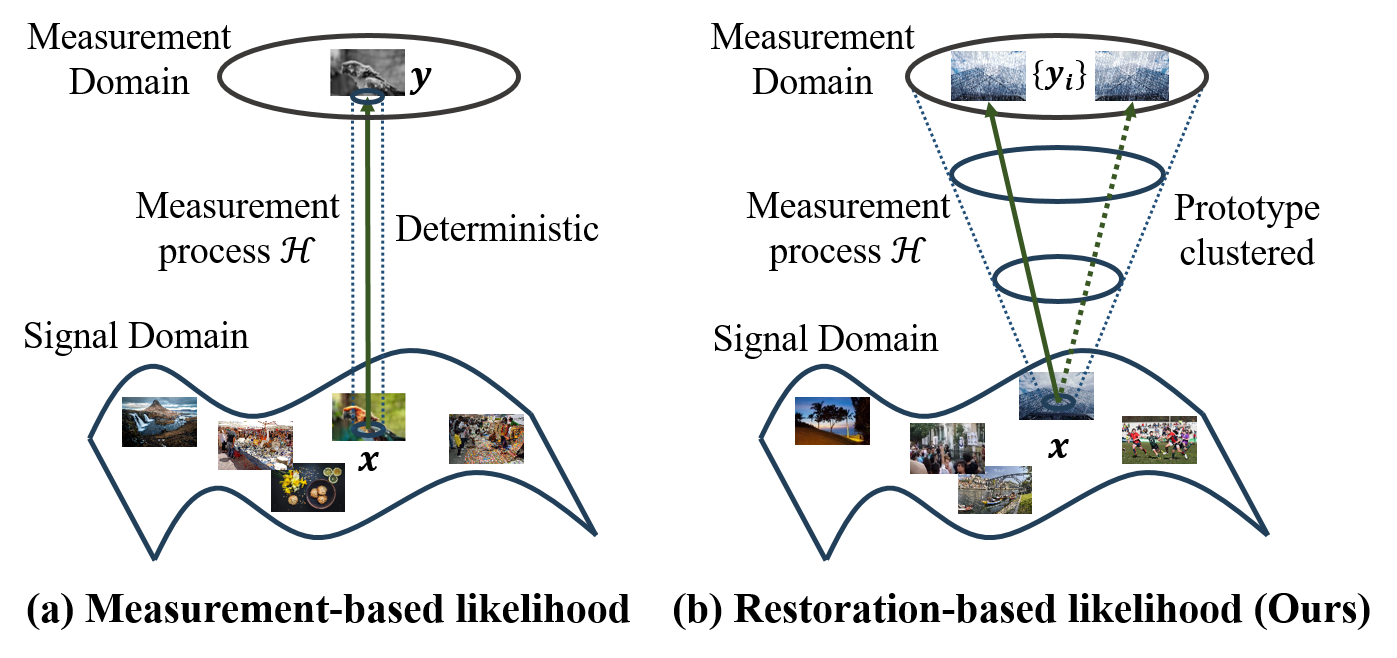}
\captionof{figure}{Visual illustration of the data consistency (likelihood) term in prevailing diffusion-based inverse problem solvers. Compared to the deterministic deterioration process of the measurement-based likelihood, the restoration-based likelihood is more adaptable for versatile inverse problems with capricious unpredictable disturbance, such as varying weather conditions and irregular manual disruption, by virtue of the augmented prototype clustered deterioration processes.}  
\label{fig:tease}
\vspace{-2em}
\end{center}
\end{figure*}

Solving inverse problems with diffusion models can be crafted in multiform frameworks. 
Bayesian approach incorporates the gradients from the measurement-based likelihood, i.e., $\nabla_{\vx}\log p(\vy|\vx)$, forming the conditional score function for posterior sampling, and the data consistency can be ensured with the dependency derived from the measurement process $\vy \simeq \mathcal{H}(\vx)$, where $\mathcal{H}$ is the measurement model. 
Representative methods~\cite{chung2022improving,chung2022diffusion,song2022pseudoinverse} 
progressively extend the diffusion solvers with linear, non-linear, or even non-differentiable measurement models for increasingly complicated inverse problems.
Beyond the Bayes' formula, there are a broad range of alternatives delivering the balance between data fidelity and realistic for solving inverse problems, such as range-null space decomposition~\cite{wang2023zeroshot} and heuristic energy function with configured properties~\cite{fei2023generative,zhao2022egsde}.
These methods can be comfortably adapted to various inverse problems without retraining the diffusion model.
However, it is noteworthy that most prevailing approaches confine to the deterministic deterioration process in virtue of the stationary parameters of the measurement model, 
regardless of handcrafted~\cite{chung2022improving,chung2022diffusion,song2022pseudoinverse,wang2023zeroshot} or parameterized by well-learned neural network~\cite{fei2023generative,zhao2022egsde}.
Consequently, the desirable deteriorations mostly involve the exact digitized corruption such as image colorization, image inpainting, and phase retrieval, regardless of capricious unpredictable disturbance in real-world sceneries, due to the incongruous likelihood dependency derived for data consistency, including but not limited to varying weather conditions~\cite{zhu2023learning} or irregular manual disruption~\cite{kohler2012recording}.

Another line of works~\cite{stevens2023removing,chung2023parallel} introduce parallel diffusion models for the signal $\vx$ and the deterioration parameters in measurement model $\mathcal{H}$, and jointly estimate their score functions for posterior sampling, which release the deficiency of the deterministic deterioration process with bestowed signal reliant flexibility.
Additionally, ~\cite{laroche2023fast} investigates to alternately estimate the measurement parameters and data distribution under the traditional iterative optimization framework in the same spirit.
Albeit the flexible generated deterioration parameters, these methods remain in the paradigm of the measurement-based likelihood, and confine to the rigid deterioration formulation determined by the measurement model, which inevitably restricts their adaptability for capricious unpredictable disturbance in real-world sceneries.
Moreover, it is worth noting that aside from the aforementioned deterministic prevalent characteristic of existing likelihood terms, 
the ancillary learning of the measurement model is necessary to be realized on-the-fly, which is substantially time-consuming and inconvenient to deploy, compared to the well-prepared diffusion models for zero-shot solver.

In this work, we extend prevailing diffusion solvers for versatile inverse problems beyond the restriction of deterministic deterioration process without any extra training.
In the context of Bayes' framework, we show that the measurement-based likelihood can be renovated with restoration-based likelihood via opposite probabilistic graphic direction, forming the reliable conditional score function for posterior sampling, in what we call \textit{restorer guidance}.
Compared with measurement-based likelihood, \textit{restorer guidance} licences the patronage of various off-the-shelf restoration models, 
and implicitly extends the strictly deterministic deterioration process in measurement-based likelihood to adaptable clustered processes with supposed restorer prototype, as shown in Fig. \ref{fig:tease}.
Therefore, the incongruous dependency between the forward deterioration process $\mathcal{H}(\vx)$ and the contaminated measurement $\vy$ can be properly resolved with reliable likelihood for comprehensive applications.
Assembled with versatile restorer prototypes optionally, we can resolve inverse problems with bunch of choices for assorted sample quality and realize the proficient deterioration control with assured realistic. 
We show that our work can be formally analogous to the transition from classifier guidance~\cite{dhariwal2021diffusion} to classifier-free guidance~\cite{ho2022classifier} in the field of inverse problem solver.
Note that our method is also compatible with other frameworks beyond Bayesian, such as range-null space decomposition~\cite{wang2023zeroshot}. 

Empirically, we demonstrate the effectiveness of our method on various challenging inverse problems with unpredictable deterioration processes, including image dehazing, rain streak removal, and motion deblurring, and show that our method is a competitive inverse problem solver with superior sample quality (Fig. \ref{fig:vC}). 
Moreover, \textit{restorer guidance} is also favourable to the out-of-distribution deterioration and proficient deterioration control.

\section{Background}
\label{sec:background}

\subsection{Score-based Diffusion Models}
Score-based diffusion models smoothly transform data distribution to spherical Gaussian distribution with a diffusion process, and reverse the process with score matching to synthesize samples.
The \textit{forward process} $\{\vx_t\}_{t \in [0, T]},\,\vx_t\in \mathbb{R}^D$, can be represented with the following It$\hat{\rm{o}}$ stochastic differential equation (SDE)~\cite{song2020score}:
\begin{equation}\
\label{eq:forward-sde}
    d\vx = \vf(\vx,t)dt + g(t)d\vw,
\end{equation}
where $\vf(\cdot,t): \mathbb{R}^D \to \mathbb{R}^D$ is the drift coefficient, $g(t) \in \mathbb{R}$ is the diffusion coefficient, and $\vw \in \mathbb{R}^D$ is the standard Wiener process (a.k.a., Brownian motion). Let $ p(\vx_t)$ denotes the marginal distribution of $\vx_t$.
The data distribution is defined when $t=0$, i.e. $\vx_0 \sim p_{\text{data}}$, and the tractable prior distribution is approximated when $t=T$,  e.g.  $\vx_T \sim \mathcal{N}(\bm{0}, \boldsymbol{I})$.
$p(\vx_t|\vx_0)$ denotes the transition kernel from $\vx_0$ to $\vx_t$. Note that we always have $p_0=p_{\text{data}}$ by forward definition~\ref{eq:forward-sde}.

Samples from $ p(\vx_t)$ can be simulated via the associated \textit{reverse-time diffusion process} of \ref{eq:forward-sde}, solving from $t=T$ to $t=0$, given by the following SDE~\cite{anderson1982reverse,song2020score}
\begin{equation}
\label{eq:reverse-sde}
    \mathrm{d} \vx = [\vf(\vx,t) - g(t)^2 \nabla_{\vx_{t}} \log p(\vx_t)]\mathrm{d}t + g(t) \mathrm{d}\bar\vw,
\end{equation}
where $\overline{\vw}$ is the reverse-time standard
Wiener process, and $\mathrm{d} t$ is an infinitesimal negative timestep.
The reverse process of \ref{eq:reverse-sde} can be derived with the \textit{score function} $\nabla_{\vx_t} \log p(\vx_t)$ at each time $t$, which is typically replaced with $\nabla_{\vx_t}\log p(\vx_t|\vx_0)$ in practice, and is approximated via score-based model $\vs_\theta(\vx_t, t)$ trained with \textit{denoising score matching objective}~\cite{vincent2011connection}:
\begin{equation}
\label{eq:dsm}
    \theta^* = \argmin_\theta \Ed_{t, \vx_t, \vx_0}\left[\|\vs_\theta(\vx_t, t) - \nabla_{\vx_t}\log p(\vx_t|\vx_0)\|_2^2\right],
\end{equation}

where $\varepsilon \simeq 0$ is a small positive constant. 
Score matching ensure the optimal solution $\theta^*$ converges to $\nabla_{\vx_t} \log p(\vx_t) \simeq \vs_{\theta^*}(\vx_t, t)$ with sufficient data and model capability.
One can replace the score function in \ref{eq:reverse-sde} with $\vs_{\theta^*}(\vx_t, t)$ to calculate the \textit{reverse-time diffusion process} and solve the trajectory with numerical samplers, such as Euler-Maruyama, Ancestral sampler~\cite{ho2020denoising}, probability flow ODE~\cite{song2020score}, DPM-Solver~\cite{lu2022dpm}, amounts to sampling from the data distribution with the goal of generative modeling.

\subsection{Solving Inverse Problem with Diffusion Models}
Solving inverse problem with diffusion model leverage the implicit prior of the underlying data distribution that the diffusion model have been learned~\cite{chung2022improving,chung2022diffusion,song2022pseudoinverse,stevens2023removing}. Formed in the Bayes' framework, we have $p(\vx|\vy) = p(\vy|\vx)p(\vx)/p(\vy)$. 
Let $\vy$ denotes the contaminated observation derived from the complete measurement $\vx$, we can straightforward modify the unconditional score function in ~\ref{eq:reverse-sde} with the following posterior formula, similar to the classifier guidance~\cite{dhariwal2021diffusion}: 
\begin{equation}
\label{eq:Bayes}
    \nabla_{\vx_t} \log p(\vx_t|\vy) = \nabla_{\vx_t}\log p(\vx_t) + \nabla_{\vx_t}\log p(\vy|\vx_t),
\end{equation}
where the prior term can be approximated via the pre-trained score model $\vs_{\theta^*}(\vx_t, t)$, and the likelihood term can be acquired via the compound of the Tweedie’s formula~\cite{efron2011tweedie} and the measurement model from $\vx$ to $\vy$ to ensure the data consistency.
Simply replacing the score function in \ref{eq:reverse-sde} with ~\ref{eq:Bayes} enable the conditional \textit{reverse-time diffusion process} for posterior sampling:
\begin{equation}
\label{eq:reverse-sde-posterior}
    \mathrm{d}\vx = \left[\vf(\vx,t) - g(t)^2\nabla_{\vx_t} \log p(\vx_t|\vy)\right]\mathrm{d}t + g(t)\mathrm{d}\bar\vw,
\end{equation}
where the first term promise the realistic powered by diffusion manifold constraint, and the second term ensure the data fidelity.
It is worth noting that the likelihood can be further approximated with heuristic energy function with configured properties~\cite{zhao2022egsde,fei2023generative}.

\section{Methods}
\label{methods}

\subsection{Measurement-based likelihood}
Recall that the posterior sampling from the conditional score function (Eq.~\ref{eq:reverse-sde-posterior}) require the likelihood term $\nabla_{\vx_t} \log p(\vy|\vx_t)$ to provide the guidance which is intractable to compute.
Pioneer works typically factorize $p(\vy|\vx_t)$ with the marginalization over $\vx_0$, considering the underlying graphic model: 
\begin{equation}
\label{eq:factorize_yxt}  
    p(\vy|\vx_t) = \int_{\vx_0} p(\vy|\vx_0, \vx_t)p(\vx_0|\vx_t) d\vx_0 = \int_{\vx_0} p(\vy|\vx_0)p(\vx_0|\vx_t) d\vx_0. 
\end{equation}
Note that $\vx_t$ is independent of the measurement $\vy$ when conditioned on $\vx_0$. In this way, we can accordingly approximating the $p(\vx_0|\vx_t)$  via one-step denoising process with Tweedie's formula~\cite{efron2011tweedie}, and solving the $p(\vy|\vx_0)$ from the measurement model. 
Unfortunately, the prevalent measurement-based likelihood is restricted to the deterministic deficiency of the measurement model, impeding the diffusion solvers for versatile inverse problems. 

\subsection{Restoration-based likelihood}
\label{sec:RG}
To address the abovementioned limitations, we show that the measurement-based likelihood can be renovated with restoration-based likelihood for data consistency, in what we call \textit{restorer guidance}. 
Compared with measurement-based likelihood, \textit{restorer guidance} licencing the patronage of various off-the-shelf restoration models for powerful diffusion solvers, considering their comprehensive sensitivity to multifarious deterioration process.
We first write the factorized restoration-based likelihood $\hat p(\vx_t|\vy)$ as the following for comparison, and the modified conditional score function together with the restorer guided posterior sampling will be introduced later.
\begin{equation}
\label{eq:factorize_xty}  
    \hat p(\vx_t|\vy) = \int_{\vx_0} p(\vx_t|\vx_0, \vy)p(\vx_0|\vy) d\vx_0 = \int_{\vx_0} p(\vx_t|\vx_0)p(\vx_0|\vy) d\vx_0, 
\end{equation}
\begin{figure}
  \vspace{-0.6cm}
  \begin{center}
  \includegraphics[width=0.52\textwidth]{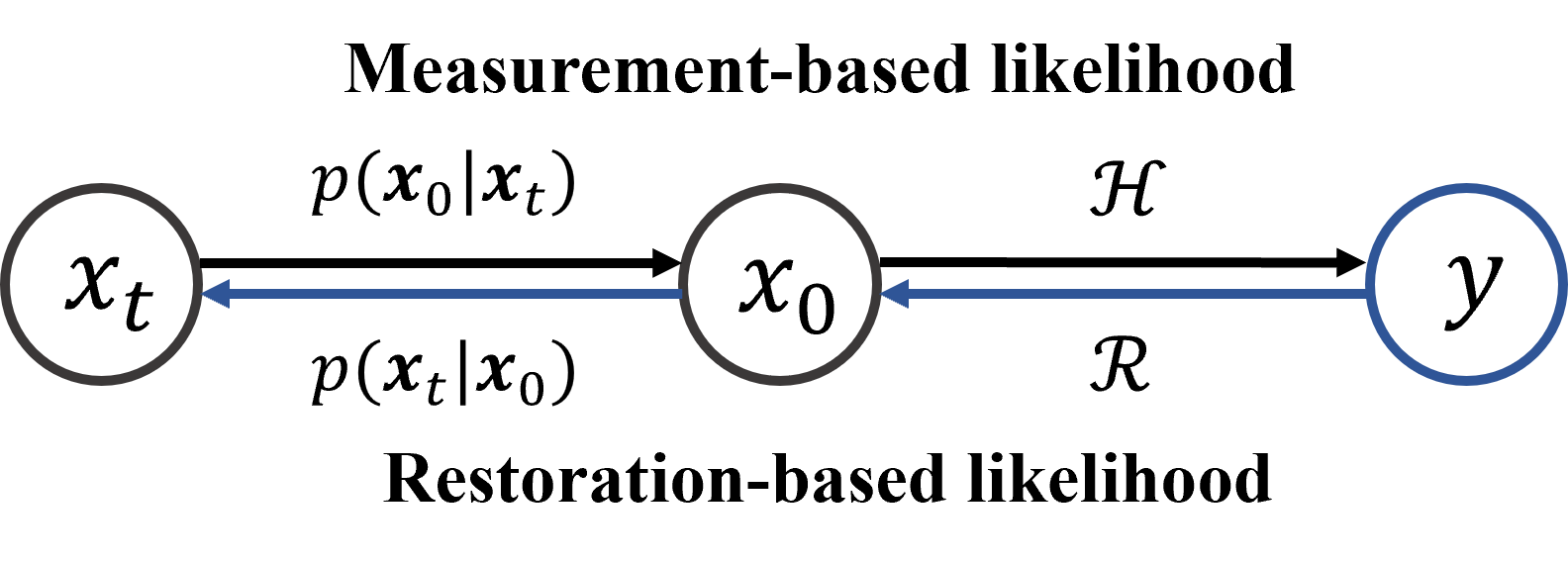}
  \vspace{-1em}
\captionof{figure}{Probabilistic graphic model of the likelihood term. The direction of restoration-based likelihood is opposite to the prevailing measurement-based likelihood.}
\label{fig:prob_graph}
\end{center}
\vspace{-1.6em}
\end{figure}
where the measurement $\vy$ is independent of $\vx_t$ when conditioned on $\vx_0$.

Note that the probabilistic graphic direction of Eq. \ref{eq:factorize_xty} is opposite to the measurement-based likelihood (Eq. \ref{eq:factorize_yxt}) for confident data consistency, as shown in Fig. \ref{fig:prob_graph}.
Solving $p(\vx_0|\vy)$ with assorted restoration models $\mathcal{R}$ enable the implicit establishment of adaptable measurement cluster processes, compared to the deterministic $p(\vy|\vx_0)$.
While the $p(\vx_t|\vx_0)$ can be directly derived from the forward process, e.g., $p(\vx_t|\vx_0) \sim \mathcal{N}(\sqrt{\bar\alpha(t)} \vx_{0}, (1-\bar\alpha(t))\boldsymbol{I})$, in the case of VP-SDE or DDPM~\cite{ho2020denoising}.
Therefore, we have $\hat p(\vx_t|\vy) \sim \mathcal{N}(\sqrt{\bar\alpha(t)} \mathcal{R}(\vy), (1-\bar\alpha(t))\boldsymbol{I})$, considering the deterministic process of $p(\vx_0|\vy)$.
The score of the restoration-based likelihood can be written as:
\begin{equation}
\label{eq:likelihood-score}
     {\nabla_{\vx_t} \log \hat p(\vx_t|\vy) \simeq - \frac{1}{\sigma^2_t} \nabla_{\vx_t} \|\vx_t - \sqrt{\bar\alpha(t)}\mathcal{R}(\vy)\|_2^2}  
\end{equation}
where $\sigma_t$ is exactly the standard deviation of $\hat p(\vx_t|\vy)$. 
Note that the underlying distribution of the mean-reverting error (Eq. \ref{eq:likelihood-score}) is regularized into the time-constant $\epsilon_t \sim \mathcal{N}(\bm{0}, \boldsymbol{I})$, and we empirically transform it to time-dependent $\epsilon_t \sim \mathcal{N}(\bm{0}, \sigma_t^2)$ by discarding the normalized term $1/\sigma_t^{2}$ for accommodative guidance that related to the noise schedule. 
Once we obtain the $\nabla_{\vx_t} \log p(\vx_t|\vy)$, we can freely plug it into the modified conditional score function for restorer guided posterior sampling, as presented in Sec.~\ref{sec:3.3}. 

\subsection{Posterior Sampling from Restorer Guidance}
\label{sec:3.3}
To enable the posterior sampling from the restoration-based likelihood and forming the renovative conditional score function, we rewrite the likelihood term in Eq. \ref{eq:Bayes} as following via Bayes' rule:
 \begin{equation}
\label{eq:Bayes2}
    \nabla_{\vx_t} \log p(\vy|\vx_t) = \nabla_{\vx_t}\log \hat p(\vx_t|\vy) - \nabla_{\vx_t}\log p(\vx_t),
\end{equation}
which decently translates the measurement-based likelihood $\nabla_{\vx_t} \log p(\vy|\vx_t)$ to restoration-based likelihood $\nabla_{\vx_t}\log \hat 
p(\vx_t|\vy)$.
Therefore, the conditional score function $\nabla_{\vx_t} \log p(\vx_t|\vy)$ can be simply accessed by plugging in the derivation from Eq. \ref{eq:Bayes2} to Eq. \ref{eq:Bayes}.
Considering the typical parameters $w$ that controls the strength of the measurement-based guidance, i.e., $w\nabla_{\vx_t} \log p(\vy|\vx_t)$, we have:
\begin{equation}
\label{eq:Bayes3}
\footnotesize
    \nabla_{\vx_t} \log p(\vx_t|\vy) = (1-w)\nabla_{\vx_t}\log p(\vx_t) + w\nabla_{\vx_t}\log \hat p(\vx_t|\vy),
\end{equation}
where $w$ is generally a positive number for smooth control between data consistency and realistic, and the restoration-based likelihood $\nabla_{\vx_t} \log \hat p(\vx_t|\vy)$ is then used in $\nabla_{\vx_t} \log p(\vx_t|\vy)$ when posterior sampling from diffusion solvers.
In the context of the restoration-based likelihood, the data consistency is further exteriorized as restorer intensity to flexibly release the power of the restoration model, which is unrealizable in the opposite measurement-based likelihood.
Substituting the derived restoration-based likelihood in Eq. \ref{eq:likelihood-score} enable the posterior sampling from the restorer guidance.
The conditional score function in Eq. \ref{eq:Bayes3} formally comes to be:
\begin{equation}
\label{eq:dps_gauss}
\footnotesize
    {\nabla_{\vx_t} \log p_t(\vx_t|\vy) \simeq  \eta \vs_{\theta^*}(\vx_t, t) -  {\rho} \nabla_{\vx_t} \|\vx_t - \sqrt{\bar\alpha(t)}\mathcal{R}(\vy)\|_2^2},  
\end{equation}
where we set the parameters $\eta$ and $\rho$ as harmonic step size for the unconditional prior term and restoration-based likelihood term, and the strict constrain in Eq. \ref{eq:Bayes3} is released, considering the unique balance between restorer intensity and data realistic countered by the diffusion model.

\textbf{Related to the classifier-free guidance.} It is worth noting that the prevailing measurement-based likelihood is homologous to the classifier guidance~\cite{dhariwal2021diffusion}, considering the same role of the classifier and the measurement model played in the conditional score function (Eq.~\ref{eq:Bayes}).
Beyond, we show that the \textit{restorer guidance} is formally analogous to the classifier-free guidance~\cite{ho2022classifier} in terms of the likelihood decomposition (Eq. \ref{eq:Bayes2}).
While the difference lies in the conditional prior term $\nabla_{\vx_t} \log p(\vx_t|\vy)$ assumed in Eq.~\ref{eq:Bayes}, instead of unconditional $\nabla_{\vx_t} \log p(\vx_t)$, resulting in:
\begin{equation}
\label{eq:classifier-free}
\footnotesize
    \nabla_{\vx_t} \log p(\vx_t|\vy) = (w+1)\nabla_{\vx_t}\log \hat p(\vx_t|\vy) - w\nabla_{\vx_t}\log p(\vx_t),
\end{equation}
which is exactly the classifier-free guidance that sampling from the linear combination of the unconditional score and conditional score estimates. 
Compared with restorer guidance, the conditional score in Eq. \ref{eq:classifier-free} is provided by extra-trained conditional diffusion model, rather than arbitrary off-the-shelf restorers.
It also explains why the constrain in Eq. \ref{eq:Bayes3} need to be released as the data realistic cannot be guaranteed by the restorer-based likelihood term, compared to the diffusion guidance.

\subsection{Extension of the Restorer Guidance}
The \textit{restorer guidance} of Eq. \ref{eq:dps_gauss} presents conceptual transition from measurement-based likelihood to restoration-based likelihood ideologically, 
and we show that it can be further extended to release the great potential of alternative restorers for constructing powerful diffusion solvers.
We here provide three major extensions for original \textit{restorer guidance} in the following.

\noindent
\textbf{Step 1: Gradient orientation.} Apart from the measurement-based likelihood that the conditional gradients from $\nabla_{\vx_t} \log p(\vy|\vx_t)$ are definitely traced back to the current $\vx_t$,
the likelihood gradients in restorer guidance $\nabla_{\vx_t} \log p(\vx_t|\vy)$ can be solely dependent on the current unconditional update, 
attributing to the opposite probabilistic graphic direction that no longer proceeding from the  $\vx_t$.
Therefore, the parallel gradient update in original conditional score function can be replaced with serial update for efficient gradient calculation.
Let $\vx'_{t-1}$ denotes the unconditional update of $\vx_t$, we can rewrite the Eq. \ref{eq:dps_gauss} as:
\begin{equation}
\label{eq:dps_ex1}
\footnotesize
    {\nabla_{\vx_t} \log p_t(\vx_t|\vy) \simeq  \eta \vs_{\theta^*}(\vx_t, t) -  {\rho} \nabla_{\vx'_{t-1}}\|\vx'_{t-1} - \sqrt{\bar\alpha(t)}\mathcal{R}(\vy)\|_2^2},  
\end{equation}
where Eq. \ref{eq:dps_ex1} can be approximately regarded as serial update for prior term and likelihood term, and the alteration of the gradient orientation avert the further step tracing derivation.

\newcommand{\DDIM}{\color{xkcdLightGrey}}
\newcommand{\base}{\color{xkcdBlack}}
\begin{figure}[t]
\begin{minipage}{.98\textwidth}
    \vspace{-0.2cm}
    \begin{algorithm}[H]
    \setstretch{1.11}
           \footnotesize
           \caption{Posterior sampling from Restorer Guidance}
           \label{alg:ddpm-sampling}
            \begin{algorithmic}[1]
             \Require $N$, $\vy$, $\eta$, $\rho$, $\zeta$, ${\{\tilde\sigma_t\}_{t=1}^N}$, $\mathcal{R}(\cdot)$
             \State {$\vx_N \sim \mathcal{N}(\sqrt{\bar\alpha_N} \vy, (1-\bar\alpha_N)\boldsymbol{I})$}
              \For{$t=N-1$ {\bfseries to} $0$}
                 \State{{$\hat\vs \gets \vs_\theta(\vx_t, t)$}}
                 \State{{$\hat\vx_{0|t} \gets \frac{1}{\sqrt{\bar\alpha_t}}(\vx_t + (1 - \bar\alpha_t)\hat\vs)$}}
                 \State{$\vz \sim \mathcal{N}(\bm{0}, \bm{I})$}
                 \State{$\vx'_{t-1} \gets \frac{\sqrt{\alpha_t}(1-\bar\alpha_{t-1})}{1 - \bar\alpha_t}\vx_t + \frac{\sqrt{\bar\alpha_{t-1}}\beta_t}{1 - \bar\alpha_t}\mathcal{R}(\hat\vx_{0|t})+   \tilde\sigma_t \vz$} 
                 \Comment{Restorer traveling}
                 \DDIM\State{$\vx'_{t-1} \gets - \sqrt{1 - \bar\alpha_t}\sqrt{1-\bar\alpha_{t-1}-\tilde\sigma_{t-1}}\hat\vs + \mathcal{R}(\hat\vx_{0|t}) + {\tilde\sigma_t \vz}$}
                 \Comment{(DDIM sampler)}
                 \base\State{$\vr_t \gets {\rho} \nabla_{\vx'_{t-1}} \|\vx'_{t-1} - \sqrt{\bar\alpha_t}\mathcal{R}(\vy)\|_2^2$}
                 \Comment{Gradient orientation}
                 \State{$\vm_t \gets {\zeta} \nabla_{\vx'_{t-1}} \|\vx'_{t-1} - \sqrt{\bar\alpha_t}\vy\|_2^2$}
                 \Comment{Measurement boosting}
                 \State{$\vx_{t-1} \gets {\eta}\vx'_{t-1} - \vr_t + \vm_t$}
                 \Comment{Posterior sampling}
              \EndFor
              \State {\bfseries return} ${\vx}_0$
            \end{algorithmic}
    \end{algorithm}
\end{minipage}
\vspace{-0.5em}
\end{figure}

\noindent
\textbf{Step 2: Restorer traveling.} The likelihood in original restorer guidance only involves $\mathcal{R}(\vy)$ for the application of the restoration model, which is insufficient to release the great potential of alternative restorers for powerful solvers.
Proceeding from this limitation, we show that the restorer prototype can be invoked recursively for optional choice, with the escort of the diffusion model.
Besides the guidance provided from restorers, we explicitly apply the restoration model on the one-step denoising result $\hat\vx_{0|t}$ for reliable data consistency, forming the unconditional update of $\vx'_{t-1}$ in the case of DDPM sampling as following:
\begin{equation}
    \vx'_{t-1} \gets \frac{\sqrt{\alpha_t}(1-\bar\alpha_{t-1})}{1 - \bar\alpha_t}\vx_t + \frac{\sqrt{\bar\alpha_{t-1}}\beta_t}{1 - \bar\alpha_t} \mathcal{R}(\hat\vx_{0|t}) +  {\tilde\sigma_t \vz},
\end{equation}
where $\boldsymbol{\vz}\sim \mathcal{N}(0,\mathbf{I})$, we denote the $\alpha(t)$ as $\alpha_t$ for simplicity, and $\beta_t \triangleq 1-\alpha_t$, $\tilde\sigma_t$ is the reverse diffusion variance.
It is worth noting that the explicit restoration of $\hat \vx_{0|t}$ will not hinder the likelihood gradients derived from the \textit{restorer guidance}, which is detached from the unconditional update $\vx'_{t-1}$ (Eq. \ref{eq:dps_ex1}). 
We provide this extension as optional and the lightweight restorer will cause negligible computational burden, compared to the unconditional score model $\vs_{\theta^*}(\vx_t, t)$.

\noindent
\textbf{Step 3: Measurement boosting.} The \textit{restorer guidance} presented so far only depends on the information provided from the restoration model, ignoring the original information possessed in the measurement $\vy$, which prone to lead the suboptimal prototype-biased solving results.
To this end, 
we reformulate the conditional score function in Eq. \ref{eq:dps_gauss} to incorporate the information across both sides of the restorer.
Combining with above two extensions, we have the following complete score function of the \textit{restorer guidance} for posterior sampling:
\begin{equation}
\footnotesize
\label{eq:dps_ex3}
    \nabla_{\vx_t} \log p_t(\vx_t|\vy) \simeq  \eta \vs_{\theta^*}(\vx_t, t) -  {{\rho} \nabla_{\vx'_{t-1}} \|\vx'_{t-1} - \sqrt{\bar\alpha_t}\mathcal{R}(\vy)\|_2^2} + {{\zeta} \nabla_{\vx'_{t-1}} \|\vx'_{t-1} - \sqrt{\bar\alpha_t}\vy\|_2^2},
\end{equation}
where $\zeta$ is a parameter that controls the strength of score derived from the measurement, $\zeta \ll \rho$, and we perform the gradient ascent in this term to boost the performance of the diffusion solver.
We provide the full version of the posterior sampling from the \textit{restorer guidance} with DDPM and DDIM sampler in Algorithm \ref{alg:ddpm-sampling}.
And the \textit{restorer guidance} in the context of the range-null space decomposition is provided in the supplementary.

\subsection{Application of the Restorer Guidance}
\label{sec:app}
The \textit{restorer guidance} release the deterministic deficiency of the measurement-based likelihood for versatile inverse problems, with the acceding of assorted restoration models considering their comprehensive sensitivity to multifarious deterioration processes.
Aside from this, we show that the \textit{restorer guidance} can further be applied to other cases with promising sample quality and advanced performance.

\noindent
\textbf{Deterioration control.} The step parameter of the restoration-based likelihood provides us the ability to flexibly control the restorer intensity with desired deterioration removal extent; see Fig. \ref{fig:Control}.
Additionally, we show that the deterioration can be further strengthened with simply reversing the gradient directions of the likelihood terms in Eq. \ref{eq:dps_ex3}, resulting in the proficient deterioration control of both sides. 
The extension of the restorer traveling will be disabled in the case of deterioration control, while the sample realistic in deterioration strengthen can be assured with the diffusion model.

\noindent
\textbf{Out-of-distribution processing.} The \textit{restorer guidance} is capable of handling out-of-distribution deterioration beyond the alternative restorer prototype.
Formally, in that case, the conditional gradients provided from the restoration-based likelihood is unreliable, on account of the unstable results of $\mathcal{R}(\vy)$.
We show that through restorer traveling and amplified measurement boosting, the performance of diffusion solvers on out-of-distribution deterioration can be significantly advanced; see Tab. \ref{Tab:OR} and \ref{Tab:OB}.

\section{Experiments}
\label{sec:experiments}
The experiments are performed to verify the behavior and potential properties of the \textit{restorer guidance}, in comparison with previous measurement-based likelihood as well as the incorporated prototypes, and extend the prevailing diffusion solvers for versatile unpredictable disturbance in real-world scenes. 

\subsection{Implementation details}
\noindent
\textbf{Tasks and Metrics.} We experimentally evaluate \textit{restorer guidance} on three  challenging inverse problems with inaccessible deterioration processes, including image dehazing, rain streak removal, and motion deblurring.
The evaluated datasets include 500 images in SOTS-Outdoor~\cite{li2018benchmarking}, 100 images in Rain100L~\cite{yang2017deep}, and 1111 images in GoPro~\cite{nah2017deep}.
We consider the following metrics including the Learned Perceptual Image Patch Similarity (LPIPS)~\cite{zhang2018unreasonable} and Fréchet Inception Distance (FID)~\cite{heusel2017gans} for perceptual measurement, and Peak Signal to Noise Ratio (PSNR) and Structural Similarity Index Measure (SSIM) for distortion evaluation.

\newcommand{\baseline}{\color{xkcdGreyishBrown}}
\begin{table*}[t]
\renewcommand\arraystretch {1.3}
\caption{
Quantitative comparison of solving versatile inverse problems with competitive solvers. 
The baseline results of restorer prototype are in {\baseline brown}. 
\textbf{Bold}: best, \underline{underline}: second best.
}
\vspace{-0.3em}
\centering
\resizebox{1.0\textwidth}{!}{
\begin{tabular}{lcccccccccccc}
\toprule
{} & \multicolumn{4}{c}{\textbf{Image Dehaze}} & \multicolumn{4}{c}{\textbf{Rain streak removal}} &
\multicolumn{4}{c}{\textbf{Motion Deblur}}\\
\cmidrule(lr){2-5}
\cmidrule(lr){6-9}
\cmidrule(lr){10-13}
{\textbf{Method}} & {PSNR $\uparrow$} & {SSIM $\uparrow$} & {FID $\downarrow$} & {LPIPS $\downarrow$}  & {PSNR $\uparrow$} & {SSIM $\uparrow$} & {FID $\downarrow$} & {LPIPS $\downarrow$} & {PSNR $\uparrow$} & {SSIM $\uparrow$} & {FID $\downarrow$} & {LPIPS $\downarrow$}\\
\midrule
NAFNet~\cite{chen2022simple}
& \baseline30.12 & \baseline0.973 & \baseline4.88 & \baseline0.015
& \baseline33.13 & \baseline0.951 & \baseline26.93 & \baseline0.079 
& \baseline33.71 & \baseline0.947& \baseline8.82 & \baseline0.078\\
MPRNet~\cite{zamir2021multi} 
& 27.33 & 0.962 & 8.46 & 0.023 
& \textbf{34.95} & \underline{0.959} & 26.86 & 0.073 
&32.66 & 0.936& 10.98 & 0.089\\
IR-SDE~\cite{luo2023image}
& 24.90 & 0.924 & 9.45 & 0.039 
& \underline{34.20} & \textbf{0.964} & \textbf{10.30} & \textbf{0.019} 
& 30.63 & 0.901 & \textbf{6.33} & \textbf{0.062}\\
DPS~\cite{chung2022diffusion} 
& 17.29 & 0.650 & 58.78 & 0.276 
& 23.18 & 0.627 & 142.55 & 0.340 
& 24.86 & 0.742 & 83.96 & 0.371\\
DDNM~\cite{wang2023zeroshot}
& 12.68 & 0.556 & 31.72 & 0.217 
& 12.96 & 0.453 & 178.24 & 0.366 
& 25.52 & 0.752 & 60.83 & 0.304\\
\cmidrule(l){1-13}
Restorer guidance - \textit{Bayesian}
&\textbf{30.21} & \textbf{0.975} & \textbf{4.58} & \textbf{0.013} 
& 33.54 & 0.957 & \underline{25.71} &\underline{0.071}
& \textbf{34.28} & \textbf{0.953} & \underline{7.59} & \underline{0.064}\\
Restorer guidance - \textit{Null-space} & \underline{30.17} & \underline{0.973} & \underline{4.71} & \underline{0.014} 
& 33.42 & 0.952 & 26.15 & 0.074 
& \underline{33.96} & \underline{0.951} & 8.23 & 0.076\\
\bottomrule
\end{tabular}
}
\label{Tab:results_RG}
\vspace{-0.6cm}
\end{table*}

\noindent
\textbf{Sampling details.} The unconditional diffusion model is publicly available that pre-trained on ImageNet of size 256 $\times$ 256 without any finetuning~\cite{dhariwal2021diffusion}.
We adopt the DDIM sampler here, and our method can be accomplished within 10 steps for gratified sample quality.
The restorer prototypes can be selected from various off-the-shelf image restoration models that pre-trained on the suggested problem-specific datasets for proficient guidance, including RESIDE-OTS~\cite{li2018benchmarking}, Rain-combine~\cite{zamir2021multi}, and GoPro~\cite{nah2017deep} in our experiments.

\noindent
\textbf{Comparative baselines.} We perform comparison with following methods: Diffusion posterior sampling (DPS)~\cite{chung2022diffusion}, denoising diffusion null-space model (DDNM) \cite{wang2023zeroshot}, Image restoration SDE (IR-SDE)~\cite{luo2023image}, NAFNet~\cite{chen2022simple}, and MPRNet~\cite{zamir2021multi}. 
NAFNet and MPRNet are general image restoration backbone, and IR-SDE is a task-specific diffusion solver.
DPS and DDNM are measurement-based diffusion solvers for posterior sampling under separate frameworks.
Considering the inherent deficiency of the inaccessible deterioration process, we parameterize the handcrafted measurement model in DPS and DDNM with network (i.e., NAFNet) for simulating the forward capricious deterioration processes~\cite{fei2023generative}, and the same network architecture is deployed as restorer prototype for comparison.

\subsection{Quantitative results}
\label{sec:quan}
We show quantitative comparison results in Tab. \ref{Tab:results_RG},
while the restorer guidance is steadily boosting the performance of the baseline restorer prototype, i.e., NAFNet, on all tasks, regardless of frameworks in Bayesian or range-null space decomposition~\cite{wang2023zeroshot}.
This is far beyond exploiting the restoration capability conserved in restorer prototype \textit{losslessly} for visual applications, but rather breaking the upper bound of the restorer for more powerful inverse problem solvers, and also validates the compatibility of the \textit{restorer guidance} with existing unconditional score model.
On the other hand, despite the impressive performance the measurement-based methods achieved in solving deterministic inverse problems, the inherent deficiency is manifested when confronted with capricious unpredictable deterioration processes.
The likelihood derived from the incongruous dependency between the measurement model $\mathcal{H}(\vx)$ and the variable contaminated measurement $\vy$ in DPS and DDNM disable the solver behavior completely, compared to the \textit{restorer guidance} which properly resolved with opposite probabilistic graphic direction of the likelihood.
Note that the Bayesian version is referred as default in the following.

\begin{figure*}[t]
    \centering
    \includegraphics[width=\textwidth]{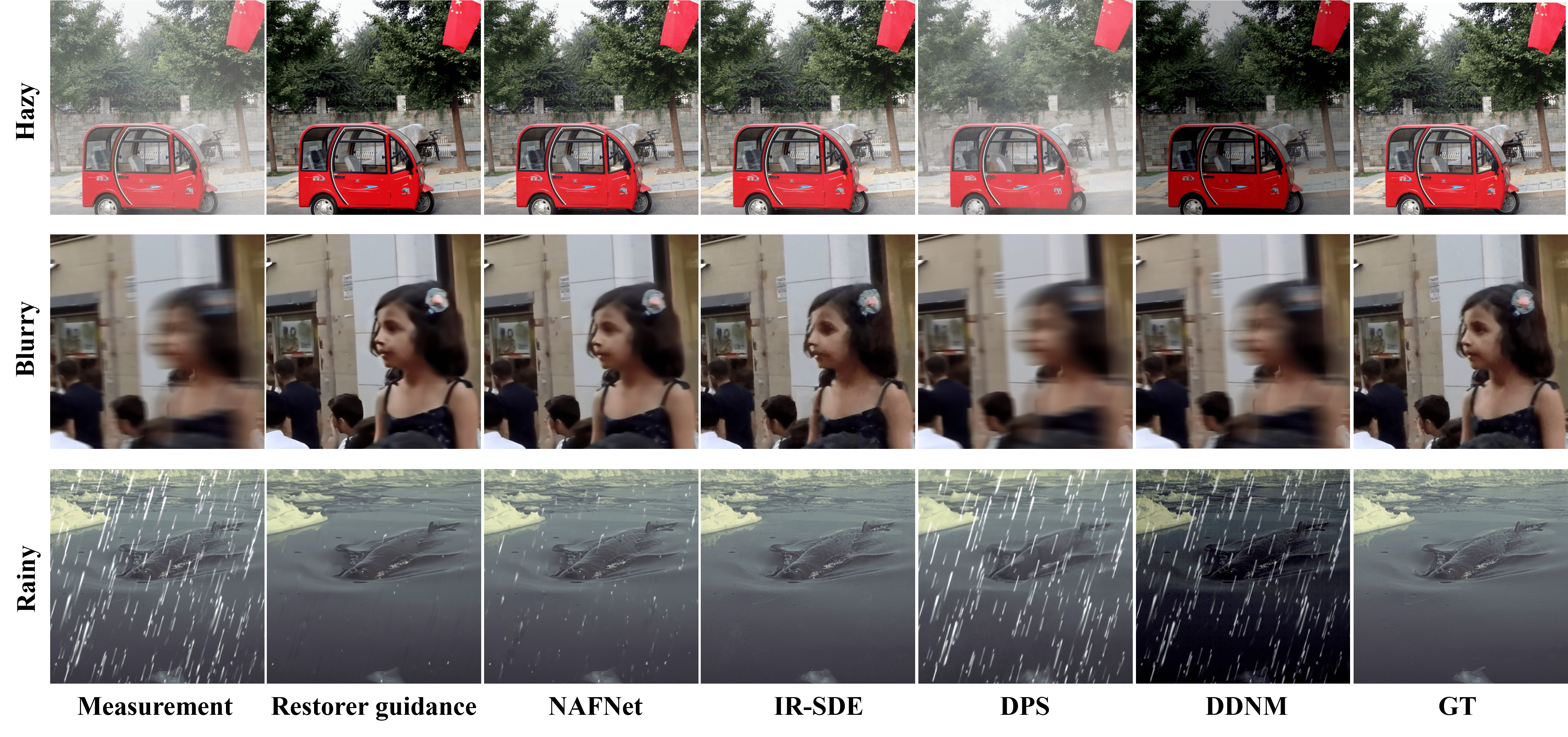}
    \vspace{-0.7cm}
    \caption{Visual comparison of restorer guidance with other inverse problem solvers on variational deterioration processes, including image dehazing, rain streak removal, and motion deblurring. The restorer prototype is deployed with NAFNet for comparison. Best viewed zoomed in.}
    \vspace{-1em}
    \label{fig:vC}  
\end{figure*}

\begin{figure}[h]
\centering
\vspace{-1em}
\begin{minipage}{0.49\linewidth}
\centering
\captionof{table}{Out-of-distribution validation of the \textit{restorer guidance}. The comparison methods are trained on Rain100L~\cite{yang2017deep} while evaluated on Rain100H~\cite{yang2017deep}.}
\renewcommand\arraystretch{1.3} {
\resizebox{\linewidth}{!}{
\begin{tabular}{lcccc}
\toprule
Methods &  PSNR$\uparrow$   & SSIM$\uparrow$& FID$\downarrow$    & LPIPS$\downarrow$ \\
\midrule
NLEDN~\cite{li2018non}  & 13.93 & 0.441& 228.5  &0.516    \\
\textit{Restorer guidance}& \cellcolor{gray!5}\textbf{16.06}  & \cellcolor{gray!5}\textbf{0.458}  & \cellcolor{gray!5}\textbf{215.2} & \cellcolor{gray!5}\textbf{0.454} \\
\cdashline{1-5}
PreNet~\cite{ren2019progressive}& 16.48&0.565&177.8     & 0.401      \\
\textit{Restorer guidance}& \cellcolor{gray!5}\textbf{19.00} & \cellcolor{gray!5}\textbf{0.587}  & \cellcolor{gray!5}\textbf{159.9} & \cellcolor{gray!5}\textbf{0.352} \\
\bottomrule
\end{tabular}}}	
\label{Tab:OR}
\end{minipage}
\hfill
\begin{minipage}{0.49\linewidth}
\centering
\captionof{table}{Out-of-distribution validation of the \textit{restorer guidance}. The comparison methods are trained on GoPro~\cite{nah2017deep} while evaluated on RealBlur-J~\cite{rim2020real}.}
\renewcommand\arraystretch{1.3} {
\resizebox{\linewidth}{!}{
\begin{tabular}{lcccc}
\toprule
Methods &  PSNR$\uparrow$   & SSIM$\uparrow$& FID$\downarrow$  & LPIPS$\downarrow$  \\
\midrule
MPRNet~\cite{zamir2021multi} & 26.46 &  0.820 & 34.26  &0.156    \\  
\textit{Restorer guidance} & \cellcolor{gray!5}\textbf{26.70}  & \cellcolor{gray!5}\textbf{0.823}  & \cellcolor{gray!5}\textbf{29.87} & \cellcolor{gray!5}\textbf{0.142} \\
\cdashline{1-5}
Restormer~\cite{zamir2022restormer}& 26.57&  0.824    &  33.08 & 0.152      \\
\textit{Restorer guidance}&  \cellcolor{gray!5}\textbf{26.74} & \cellcolor{gray!5}\textbf{0.826}  & \cellcolor{gray!5}\textbf{29.65} & \cellcolor{gray!5}\textbf{0.143} \\
\bottomrule
\end{tabular}}}
\label{Tab:OB}
\end{minipage}
\vspace{-0.6em}
\end{figure}

As presented in Sec.~\ref{sec:app}, the \textit{restorer guidance} is capable of handling out-of-distribution deterioration beyond the incorporated restorer prototypes. We present the results of out-of-distribution validation in Tab.~\ref{Tab:OR} and \ref{Tab:OB} for rain streak removal and motion deblurring, respectively. While the result for image dehazing can be found in the supplementary due to the limited space.
In Tab.~\ref{Tab:OR}, the comparison methods are trained on Rain100L~\cite{yang2017deep} while evaluated on Rain100H~\cite{yang2017deep}, differing from the deterioration strength.
In Tab.~\ref{Tab:OB}, the comparison methods are trained on GoPro~\cite{nah2017deep} while evaluated on RealBlur-J~\cite{rim2020real}, differing from the underlying deterioration prototype.
Observing that the \textit{restorer guidance} is expert at deterioration within the prototype clustered process of the restorer, while releasing the scope of the deterioration strength (Tab.~\ref{Tab:OR}). 
Moreover, deteriorations beyond the supposed prototype processes can also be handled well (Tab.~\ref{Tab:OB}), with relatively modest improvement compared to the strength variation.
Generally, \textit{restorer guidance} extends the deterministic deterioration process to a cluster of deterioration processes with supposed prototype of the restorer, and enables the sustained release of the restorer capability for augmented clustered space.

\begin{figure}[t]
  \begin{center}
  \includegraphics[width=0.48\textwidth]{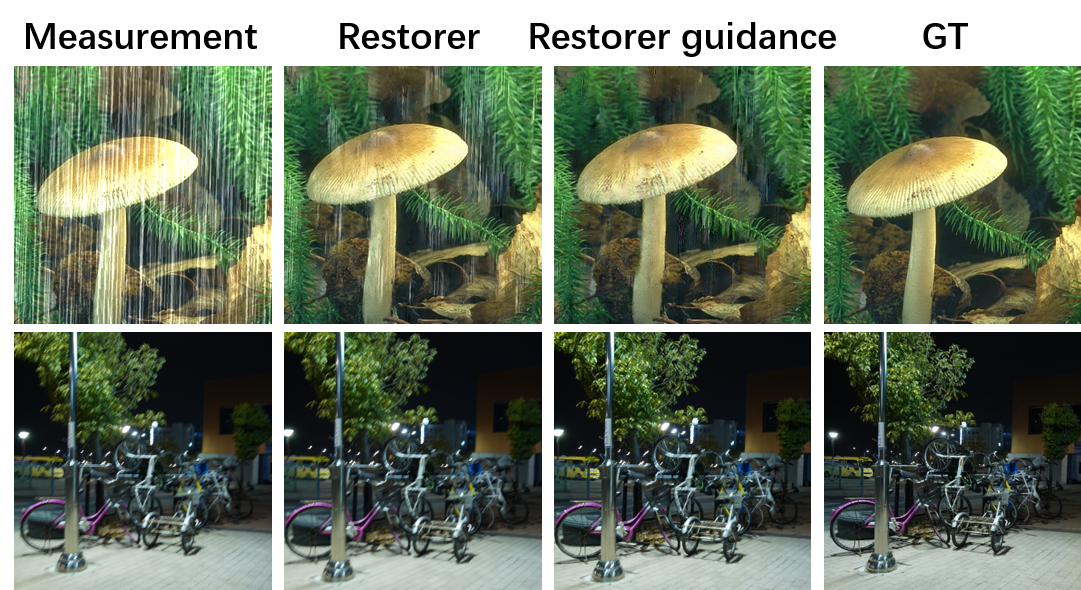}
  \vspace{-0.8em}
\captionof{figure}{Visual results of out-of-distribution validation of the restorer guidance. First row: Trained on Rain100L and evaluated on Rain100H with PreNet prototype. Second row: Trained on GoPro and evaluated on RealBlur-J with Restormer prototype.}
\label{fig:OFig}
\end{center}
\vspace{-2em}
\end{figure}

\subsection{Qualitative results and visual applications}
We provide the visual comparison in Fig.~\ref{fig:vC} to validate the effectiveness and peculiarity of the \textit{restorer guidance} qualitatively. 
Compared to the baseline restorer, \ie, NAFNet, the \textit{restorer guidance} has following \textbf{merits}: (i) Rendering the reconstructed sample with visual pleasing sample quality (e.g., red tricycle), ascribing to the unconditional score model.
(ii) Endowing the restoration process with generation capacity that synthesis the nebulous region heuristically (e.g., girl's eye).
(iii) Liberating the capability of the restorer continuously for obstinate deterioration (e.g., rain streaks) with ensured data realistic.
Compared to measurement-based solvers, the \textit{restorer guidance} is capable to provide more reliable likelihood guidance in capricious unpredictable deterioration process.

\begin{figure}[t]
  \begin{center}
  \vspace{-0.2em}
  \includegraphics[width=0.74\textwidth]{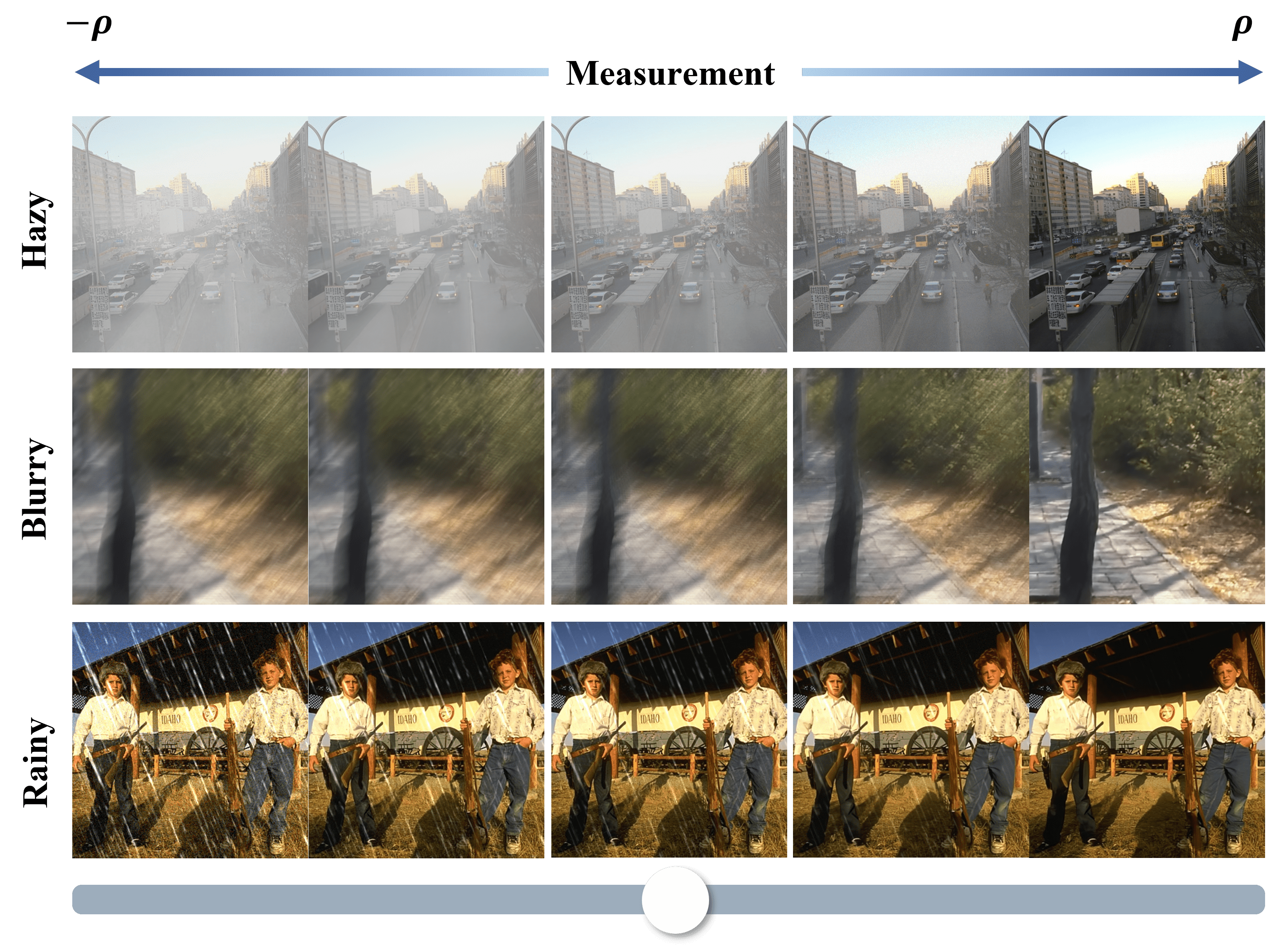}
  \vspace{-1em}
\captionof{figure}{\textit{Restorer guidance} provides us a workbench to fabricate the restoration process more flexible and controllable, with proficient deterioration expertise preserved in restorer rather than obstreperous black-box mapping.}
\label{fig:Control}
\end{center}
\vspace{-2.2em}
\end{figure}

In Fig. \ref{fig:OFig}, we provide the visual comparison of \textit{restorer guidance} on out-of-distribution deterioration. The comparison methods are exemplified as PreNet~\cite{ren2019progressive} for rain streak removal and Restormer~\cite{zamir2022restormer} for motion deblurring.
The samples drawn from the restorer guidance exhibit the greater robustness to out-of-distribution deterioration, compared to the solitary restorer prototypes.  
The proficient deterioration control achieved by \textit{restorer guidance} is shown in Fig. \ref{fig:Control}.
While one can smoothly controls the restorer intensity via the harmonic step size for desired deterioration extent, and even reverses the restoration process for amplified deterioration.
This also provides another perspective for constructing the adaptable measurement model with reversed restorers rather than opposite probabilistic graphic direction.
Generally, \textit{restorer guidance} provides us a workbench to fabricate the restoration process more flexiblely.

\subsection{Ablation studies}
We present the ablation experiments to validate the effectiveness of the suggested extensions attached to the \textit{restorer guidance}. 
The ablations are performed on problems of rain streak removal and motion deblurring, with reported PSNR and FID metrics.
In Tab. \ref{Tab:Abl}, we can see that \textit{restorer guidance} attached with extensions further bursts the potential for powerful inverse problem solvers, which is also the key to break the upper bound of the incorporated restorer prototype.
Note that the extension of the gradient orientation is adopted as default option to enable the restorer traveling and the efficient posterior sampling.

\subsection{Limitation and Discussion}
Despite the competitive performance and delightful convenience achieved by \textit{restorer guidance}, it highly depends on the capability of the alternative restorer prototype as baseline clustered deterioration processes, which prone to lead the suboptimal prototype-biased solving results. 
Beyond that, it supposed to incorporate miscellaneous restorer prototypes efficiently for allocated inaccessible deterioration to construct the unbiased \textit{restorer guidance} and release the strong dependency from the single prototype in the future.
Moreover, \textit{restorer guidance} provides us a workbench to fabricate the restoration process more flexible and controllable with proficient deterioration knowledge, and it supposed to accomplish the interconnected deterioration manipulation with discretionary user inclination in the future.

\begin{table}[t]
\renewcommand\arraystretch {1}
\caption{
Ablation experiments on major extensions attached to the restorer guidance. \textbf{RT.}: Restorer traveling. \textbf{MB.}: Measurement boosting.}
\vspace{-0.6em}
\centering
\resizebox{0.54\linewidth}{!}{
\begin{tabular}{cccccc}
\toprule
{} & {} & \multicolumn{2}{c}{\textbf{Rain streak removal}} &
\multicolumn{2}{c}{\textbf{Motion Deblur}}\\
\cmidrule(lr){3-4}
\cmidrule(lr){5-6}
\textbf{RT.} & \textbf{MB.} & {PSNR $\uparrow$} & {FID $\downarrow$} & {PSNR $\uparrow$} & {FID $\downarrow$} \\
\midrule
\ding{55} & \ding{55} & 33.06 & 26.98 & 33.67 & 8.91\\
\ding{51} & \ding{55} & 33.42 & 26.17 & 34.06 & 7.96 \\
\ding{55} & \ding{51}& 33.27 & 26.68 & 33.84 & 8.62 \\
\ding{51} & \ding{51} & \textbf{33.54} & \textbf{25.71} & \textbf{34.28} & \textbf{7.59}\\
\bottomrule
\end{tabular}
}
\vspace{-0.8em}
\label{Tab:Abl}
\end{table}

\section{Related work}
\label{sec: Related}
Image restoration is the classical inverse problem with nondeterministic degradation process imposed on the complete signal, reversing the process with contaminated measurement poses challenges for the solver.
Traditional methods incorporated various natural image priors to regularize the underlying solution space, including but not limited to sparse and low-rank prior~\cite{lefkimmiatis2023learning}, dark channel prior~\cite{he2010single}, and deep generative priors~\cite{pan2021exploiting,ulyanov2018deep}.
There methods confined to the deficiency of characterizing the natural image distribution comprehensively, and often resolve the inverse problem with insufficient regularization.   

Since Sohl-Dickstein et al.~\cite{sohl2015deep} modeling the intricate data distribution with inspired non-equilibrium thermodynamics, two successful classes of probabilistic generative models, denoising diffusion probabilistic models (DDPMs)~\cite{ho2020denoising} and score matching with Langevin dynamics (SMLDs)~\cite{song2019generative} have been innovatively developed, which gradually perturb data with noise until tractable distribution and reverse the process with score matching or noise prediction for sampling.
Song et al.~\cite{song2020score} amalgamates above two paradigms into a continuous generalized framework with stochastic differential equations.
Aside from various generative applications, diffusion models have also been widely appreciated in solving inverse problems.
The supervised works typically run the diffusion in the efficient space for deterioration modeling and efficient sampling, including residual space~\cite{luo2023image,yue2023resshift}, frequency space~\cite{cao2022high}, and latent space~\cite{xia2023diffir}. 
Another line of works adopt diffusion models as regularized priors for zero-shot problem solving, and inject the likelihood for conditional posterior sampling.
Pioneer works~\cite{chung2022improving,chung2022diffusion,song2022pseudoinverse} embrace the Bayes' framework and construct the measurement-based likelihood or generative-based likelihood~\cite{chung2023parallel,stevens2023removing} for data consistency.
Beyond that, \cite{wang2023zeroshot} leverage the framework of range-null space decomposition to deliver the balance between realistic and data consistency.
However, these methods are confined to the deterministic deterioration process characterized by the measurement model, and impotent to capricious unpredictable disturbance in real-world sceneries.

\section{Conclusion}
\label{sec: Conclusion}
In this work, we proposed the \textit{restorer guidance} for solving versatile inverse problems with unpredictable deterioration process, and shown that the measurement-based likelihood can be renovated with restoration-based likelihood via the opposite probabilistic graphic direction.
The \textit{restorer guidance} licencing the patronage of various off-the-shelf restoration models for powerful inverse problem solvers, 
extending the strictly deterministic deterioration process to adaptable prototype clustered processes, while attached with extensions further release the great potential of our method.
We show that our work is theoretically analogous to the transition from the classifier guidance to classifier-free guidance in the field of inverse problem solver.
Extensive experiments illustrate the effectiveness of the \textit{restorer guidance}, which is also favourable to 
the out-of-distribution deterioration and proficient deterioration control.
Code will be available soon.

\clearpage  

%
%
\bibliographystyle{splncs04}
\bibliography{main}

\clearpage  
\appendix

\section{The score of the restoration-based likelihood}
\label{sec:A1}
We here provide another perspective of the score built upon the restoration-based likelihood, which has been presented in Sec. 3.2 of the manuscript.
Considering the inevitable deviation between $\vx_0$ and $\mathcal{R}(y)$, which is so-called restorer bias, the $p(\vx_0|\vy)$ can be approximated with the following Gaussian:
\begin{equation}
    p(\vx_0|\vy) \sim \mathcal{N}(\mathcal{R}(\vy), \boldsymbol{I}),
\end{equation}
where the mean is obtained from the $\mathcal{R}(\vy)$, with the assumption that the underlying distribution of the mean-reverting error between $\vx_0$ and $\mathcal{R}(\vy)$ follows the Normal Gaussian, which is exactly the training objective of the restorer $\mathcal{R}$.
While the $p(\vx_t|\vx_0)$ can be derived from the forward process, which is a linear transform on $\vx_0$ and adds independent Gaussion noise:
\begin{equation}
\label{eq:forward-diffusion}
    \vx_t = \sqrt{\bar\alpha(t)} \vx_{0} + \sqrt{1-\bar\alpha(t)}\epsilon, \qquad  \epsilon \sim \mathcal{N}(\textbf{0},\textbf{\textit{I}}). 
\end{equation}
Thus, we have the following approximation to the score, with the consideration that the linear transformation of a Gaussian distribution is still following Gaussian,
\begin{equation}
    p(\vx_t|\vy) \sim \mathcal{N}(\sqrt{\bar\alpha(t)}\mathcal{R}(y), \sqrt{\bar\alpha(t)}\sqrt{\bar\alpha(t)}\boldsymbol{I}+\sqrt{1-\bar\alpha(t)}\textbf{\textit{I}}),
\end{equation}
which can be simplified as: 
\begin{equation}
    p(\vx_t|\vy) \sim \mathcal{N}(\sqrt{\bar\alpha(t)}\mathcal{R}(y), \textbf{\textit{I}}).
\end{equation}
Note that the above formulation of the $p(\vx_t|\vy)$ is exactly what we derived in Sec. 3.2 of the manuscript, and we here provided another perspective with the broken assumption of the deterministic $p(\vx_0|\vy)$. The score of the restoration-based likelihood can be written as the following:
\begin{equation}
\label{eq:app-likelihood-score}
     {\nabla_{\vx_t} \log p(\vx_t|\vy) \simeq - \nabla_{\vx_t} \|\vx_t - \sqrt{\bar\alpha(t)}\mathcal{R}(\vy)\|_2^2}, 
\end{equation}
where the same formulation of the likelihood score function is derived, as presented in Sec. 3.2 of the manuscript.

\section{Restorer guidance in range-null space decomposition}
\label{sec:NRSD}
Denoising diffusion null-space model (DDNM)~\cite{wang2023zeroshot} leverages the framework of range-null space decomposition to delivering the balance between the data consistency and realistic.
Considering the noise-free inverse problems first: 
\begin{equation}
\label{eq:app-21}
     \vy = \mathbf{H}\vx, 
\end{equation}
where, $\vy$ is the contaminated measurement, $\vx$ is the original complete signal, $\mathbf{H}$ is the forward measurement model. We adopt the same notations as
the what we presented in \textit{restorer guidance} for convenient comparison.
Essentially, the range-null space decomposition presented that any complete signal $\vx$ can be decomposed into two portions as following, according to the measurement model $\mathbf{H}$:
\begin{equation}
\vx\equiv\mathbf{H^{\dagger}}\mathbf{H}\vx + (\mathbf{I} - \mathbf{H^{\dagger}}\mathbf{H})\vx.
    \label{eq:rnd}
\end{equation}
where $\mathbf{H^{\dagger}}$ denotes the pseudo-inverse of $\mathbf{H}$, and $\mathbf{I}$ is the identity matrix.
Theoretically, the first part is in the range-space of $\mathbf{H}$ that responsible for the data consistency, and the second part is in the null-space of $\mathbf{H}$ that responsible for the data realistic, considering the following formula:
\begin{equation}
\mathbf{H}\vx\equiv\mathbf{H}\mathbf{H^{\dagger}}\mathbf{H}\vx + \mathbf{H}(\mathbf{I} - \mathbf{H^{\dagger}}\mathbf{H})\vx
    \equiv \mathbf{H}\vx + \mathbf{0}   \equiv \vy.
\end{equation}
While we can further simplify the formulation in Eq. \ref{eq:rnd} with Eq. \ref{eq:app-21} as:
\begin{equation}
\vx\equiv\mathbf{H^{\dagger}}\vy + (\mathbf{I} - \mathbf{H^{\dagger}}\mathbf{H})\vx.
    \label{eq:rnd2}
\end{equation}
To solve inverse problems with diffusion models, DDNM performs the above decomposition on the one-step denosing result $\hat{\vx}_{0|t}$ to enforce the data consistency on range-space and correct the harmonic data realistic on null-space, given by the following formation:
\begin{equation}
    \hat{\vx}_{0|t}=\mathbf{H^{\dagger}}\vy + (\mathbf{I} - \mathbf{H^{\dagger}}\mathbf{H})\vx_{0|t},
    \label{eq:ddnm}
\end{equation}
where $\vx_{0|t}$ is obtained from the Tweedie’s formula. Thereby, Eq.~\ref{eq:ddnm} delivers the delighted balance between data consistency and data realistic in solving inverse problems.

Directly applying Eq.~\ref{eq:ddnm} to noisy inverse problem leads to the inferior performance, due to the incongruous signal formation with unexpected Gaussian noise $\rvn$, i.e., $\vy = \mathbf{H}\vx +\rvn$.
The incongruous phenomenon can be formulated as following:
\begin{equation}
    \hat\vx_{0|t} = \mathbf{H^{\dagger}}\vy + (\mathbf{I} - \mathbf{H^{\dagger}}\mathbf{H})\vx_{0|t}=\vx_{0|t} - \mathbf{H}^{\dagger}(\mathbf{H}\vx_{0|t}-\mathbf{H}\vx) + \mathbf{H}^{\dagger}\rvn,
    \label{eq:noisy_1}
\end{equation}
where $\mathbf{H}^{\dagger}\rvn$ is the extra noise introduced in $\mathbf{\hat{x}}_{0|t}$, which is undesirable. To this end, DDNM+ modify the Eq. \ref{eq:noisy_1} as following to enforce the constrain on the decomposition:
\begin{equation}
    \hat\vx_{0|t} = \vx_{0|t} - {\mathbf{\Sigma}_{t}}\mathbf{H}^{\dagger}(\mathbf{H}\vx_{0|t}-\vy),
    \label{eq:ndm+ core}
\end{equation}
where $\mathbf{\Sigma}_{t}$ is set as step size for range-space correction to ensure the data consistency.
It is noted that Eq. \ref{eq:ndm+ core} falls into the the similar formation as measurement-based likelihood in Bayes' framework, which inevitable suffer from the deficiency of the deterministic deterioration process of the measurement model $\mathbf{H}$, and unable to handle versatile inverse problems.

\begin{figure}[t]
\begin{minipage}{.98\textwidth}
    \vspace{-0.7cm}
    \begin{algorithm}[H]
    \setstretch{1}
            \small
           \caption{Posterior sampling from Restorer Guidance - \textit{Null-space}}
           \label{alg:ddpm-ddnm}
            \begin{algorithmic}[1]
             \Require $N$, $\vy$, 
             $\{\mathbf{\Sigma}_{t}\}_{t=1}^N$, 
             $\{\mathbf{\Phi}_{t}\}_{t=1}^N$,
             $\mathcal{R}(\cdot)$
             \State {$\vx_N \sim \mathcal{N}(\sqrt{\bar\alpha_N} \vy, (1-\bar\alpha_N)\boldsymbol{I})$}
              \For{$t=N-1$ {\bfseries to} $0$}
                 \State{{$\hat\vs \gets \vs_\theta(\vx_t, t)$}}
                 \State{{$\vx_{0|t} \gets \frac{1}{\sqrt{\bar\alpha_t}}(\vx_t + (1 - \bar\alpha_t)\hat\vs)$}}
                 \State$\hat\vx_{0|t} = \vx_{0|t} - {\mathbf{\Sigma}_{t}}(\vx_{0|t}-\mathbf{R}\vy)$
                 \State{$\vz \sim \mathcal{N}(\bm{0}, \bm{I})$}
                 \State{$\vx_{t-1} \gets \frac{\sqrt{\alpha_t}(1-\bar\alpha_{t-1})}{1 - \bar\alpha_t}\vx_t + \frac{\sqrt{\bar\alpha_{t-1}}\beta_t}{1 - \bar\alpha_t}\hat\vx_{0|t} +  {\mathbf{\Phi}_{t} \vz}$} 
                 \DDIM\State{{$\vx_{t-1} \gets \hat\vx_{0|t} + {\mathbf{\Phi}_{t} \vz} - \sqrt{1 - \bar\alpha_t}\sqrt{1-\bar\alpha_{t-1}-\tilde\sigma_{t-1}}\hat\vs$}}
                 \Comment{(DDIM sampler)}
              \base\EndFor
              \State {\bfseries return} ${\vx}_0$
            \end{algorithmic}
    \end{algorithm}
\end{minipage}
\vspace{-0.5em}
\end{figure}

Considering the restoration model as $\mathbf{R}$, which is supposed to be the exact pseudo-inverse of the underlying capricious measurement model $\mathbf{H}$, and vice versa.
Therefore, we can rewrite the Eq. \ref{eq:ndm+ core} as following without any bells and whistles:
\begin{equation}
    \hat\vx_{0|t} = \vx_{0|t} - {\mathbf{\Sigma}_{t}}(\mathbf{R}\mathbf{R}^{\dagger}\vx_{0|t}-\mathbf{R}\vy).
    \label{eq:ndm+ core2}
\end{equation}
Exploiting the pseudo-inverse trick where $\mathbf{R}\mathbf{R}^{\dagger}\vx_{0|t}\cong
\vx_{0|t}$, we have the following formation:
\begin{equation}
    \hat\vx_{0|t} = \vx_{0|t} - {\mathbf{\Sigma}_{t}}(\vx_{0|t}-\mathbf{R}\vy),
    \label{eq:ndm+ core3}
\end{equation}
which is exactly the principle formation of the \textit{restorer guidance} in range-null space decomposition framework. 
We provide the complete sampling scheme of \textit{restorer guidance} applied in range-null space decomposition with DDPM sampler and DDIM sampler in Algorithm \ref{alg:ddpm-ddnm}.

\begin{figure*}[t]
    \begin{minipage}{0.52\linewidth}
    \centering
    \captionof{table}{Quantitative results of out-of-distribution validation of the \textit{restorer guidance}. The comparison methods are trained on RESIDE-OTS~\cite{li2018benchmarking} while evaluated on NH-Haze~\cite{ancuti2020nh}.}
    \renewcommand\arraystretch{1.2} {
    \resizebox{\linewidth}{!}{
    \begin{tabular}{lcccc}
    \toprule
    Methods &  PSNR$\uparrow$   & SSIM$\uparrow$& FID$\downarrow$  & LPIPS$\downarrow$  \\
    \midrule
    MSBDN~\cite{dong2020multi} & 12.76 &  0.448 & 299.6  &0.549    \\ 
    \textit{Restorer guidance} & \cellcolor{gray!5}\textbf{12.95}  & \cellcolor{gray!5}\textbf{0.451}  & \cellcolor{gray!5}\textbf{291.3} & \cellcolor{gray!5}\textbf{0.545} \\
    \cdashline{1-5}
    FFANet~\cite{qin2020ffa}& 12.06       &  0.423     &  296.1 & 0.565      \\
    \textit{Restorer guidance}&  \cellcolor{gray!5}\textbf{12.36} & \cellcolor{gray!5}\textbf{0.433}  & \cellcolor{gray!5}\textbf{292.6} & \cellcolor{gray!5}\textbf{0.553} \\
    \bottomrule
    \end{tabular}}}
    \label{Tab:OH}
    \end{minipage}
    \hspace{2mm}
    \begin{minipage}{0.45\linewidth}
    \vspace{0.4em}
        \centering
        \begin{center}
        \captionof{figure}{Visual results of out-of-distribution validation of the restorer guidance on NH-Haze dataset. First row: MSBDN prototype. Second row: FFANet prototype.}
        \vspace{-1.3em}
        \includegraphics[width=\textwidth]{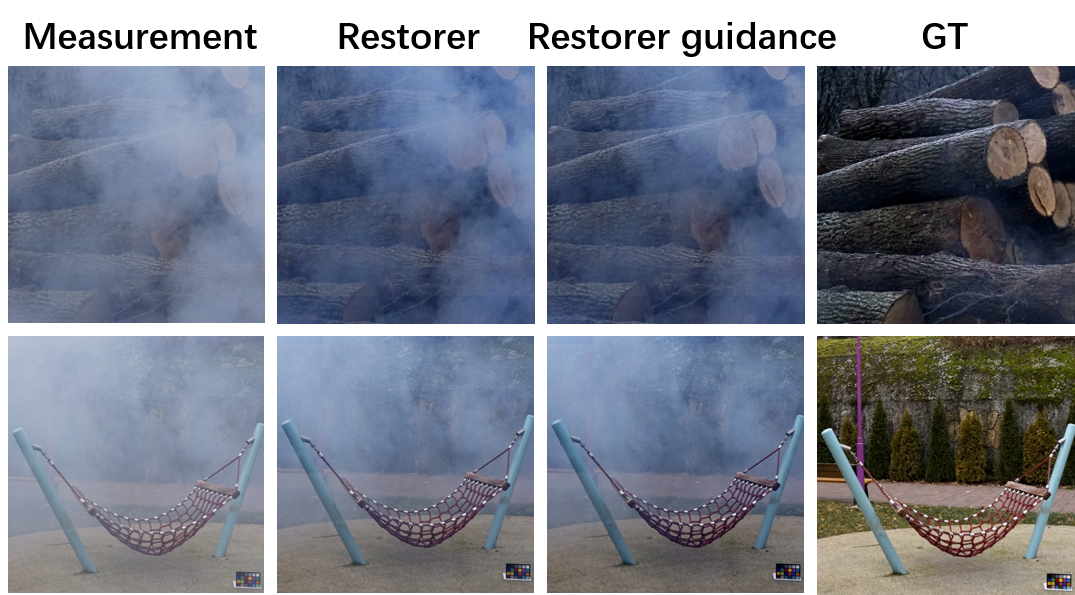}
        \vspace{-0.7em}
        \label{fig:OHFig}
\end{center}
    \end{minipage}
\vspace{-0.8em}
\end{figure*}

\section{More validation of out-of-distribution deterioration}
\label{sec:our-res}
We present the quantitative results of out-of-distribution validation in Tab. \ref{Tab:OH} for image dehazing. The comparison methods are trained on RESIDE-OTS~\cite{li2018benchmarking} and evaluated on NH-Haze~\cite{ancuti2020nh}, which are significantly differing from the underlying deterioration prototype. 
While the observation is consistent with the conclusion derived from the Sec. 4.1 of the manuscript.
The \textit{restorer guidance} handles the deterioration beyond the cluster process of the restorer prototype with relatively modest improvement, compared to the deterioration strength variation.
Fig. \ref{fig:OHFig} presents the qualitative results of the out-of-distribution validation on NH-Haze dataset, where the first row is MSBDN prototype and the second row is FFANet prototype.
The \textit{restorer guidance} has noticeable improvement compared to the incorporated restorer prototype, albeit the deterioration is far beyond the clustered processes.

\section{Detailed comparison of various measurement models}
\label{sec:measurement}
\begin{table}[b]
\vspace{-0.5em}
\setlength{\tabcolsep}{1mm}
\caption{Comparison of different measurement models $\mathcal{H}$ and the restorer guidance. The expression involved in the likelihood term are provided. $*_{\vs_\theta}$ denotes the deterioration parameters are estimated by generative score model. $*_{nn}$ denotes the deterioration process is formulated by network. $x_{pro}$ represents the prototype provided by the restorer.}
\vspace{.1cm}
\label{Tab:mea-res}
\centering
\footnotesize
\renewcommand\arraystretch{1}
\begin{tabular}{lcccc} 
\toprule
Measurement model&Likelihood term&Open formula&Adaptability& Training free\\
\midrule
Handcrafted $\mathcal{H}$~\cite{chung2022diffusion}& $y\simeq\mathcal{H}(\vx;\varphi)$&\ding{55}&\ding{55}&\ding{51}\\
Generated $\mathcal{H}$~\cite{stevens2023removing}     &$y\simeq\mathcal{H}(\vx;\varphi_{\vs_\theta}) + \vn_{\vs_\theta}$&\ding{55}&\ding{51}&\ding{55}\\
Parameterized $\mathcal{H}$~\cite{fei2023generative}&$y\simeq\mathcal{H}_{nn}(\vx;\theta)$&\ding{51}&\ding{55}&\ding{55}\\
\midrule
\textit{Restorer guidance}&$x_{pro}\simeq\mathcal{R}_{nn}(\vy;\theta)$&\ding{51}&\ding{51}&\ding{51}\\    
\bottomrule
\end{tabular}
\vspace{-1.2em}
\end{table}
We provide the detailed comparison of various measurement models and the \textit{restorer guidance} in Tab. \ref{Tab:mea-res}.
Considering the measurement model $\mathcal{H}$ with parameters $\phi$, the forward measurement process to the signal $\vx$ can be formulates as $\vy=\mathcal{H}(\vx;\phi)$, where $\vy$ is the contaminated measurement.
Typically, the formulation of the measurement model determines the profile of the measurement process, and the parameters enable the variability in the surrounding.
\textbf{i).} Handcrafted measurement model~
\cite{chung2022improving,chung2022diffusion,song2022pseudoinverse} restrict to the rigid formulation of $\mathcal{H}$, i.e., the measurement process from $\vx$ to $\vy$ is assumed to be the fixed formation such as convolution, addition, and multiplication, \etc.
While the stationary measurement parameters further restrict the measurement model to the deterministic forward process without any adaptability.
Generated measurement model~
\textbf{ii).} \cite{chung2023parallel,stevens2023removing} remain in the constrain of the rigid formulation of $\mathcal{H}$ with fixed measurement formation, however, the measurement parameters can be jointly estimated in the sampling process of signal from the generative score model, endowing the adaptability for handling capricious unpredictable measurement process.
\textbf{iii).} Parameterized measurement model~\cite{fei2023generative} extend the handcrafted measurement model with neural networks, breaking the rigid formulation of the measurement process. Albeit the learnable parameters for measurement model, the relaxation for the adaptability is still restricted, since the non-trivial implementation of the one-to-many mapping, considering the ill-posed peculiarity of the versatile inverse problems with unpredictable deterioration process.
\textbf{iv).} \textit{Restorer guidance} resolves above obstacles with opposite probabilistic graphic direction of the likelihood, compared to prevailing measurement-based methods.
The restorer prototype implicitly enable a cluster of measurement processes with desired adaptability,
rather than strict deterministic forward process.
Note that except for the handcrafted $\mathcal{H}$, both generated $\mathcal{H}$ and parameterized $\mathcal{H}$ require the extra training of the coupled measurement model, which is time-consuming and inconvenient.

\section{Additional visual results}
\begin{figure}[h]
    \centering
    \includegraphics[width=\textwidth]{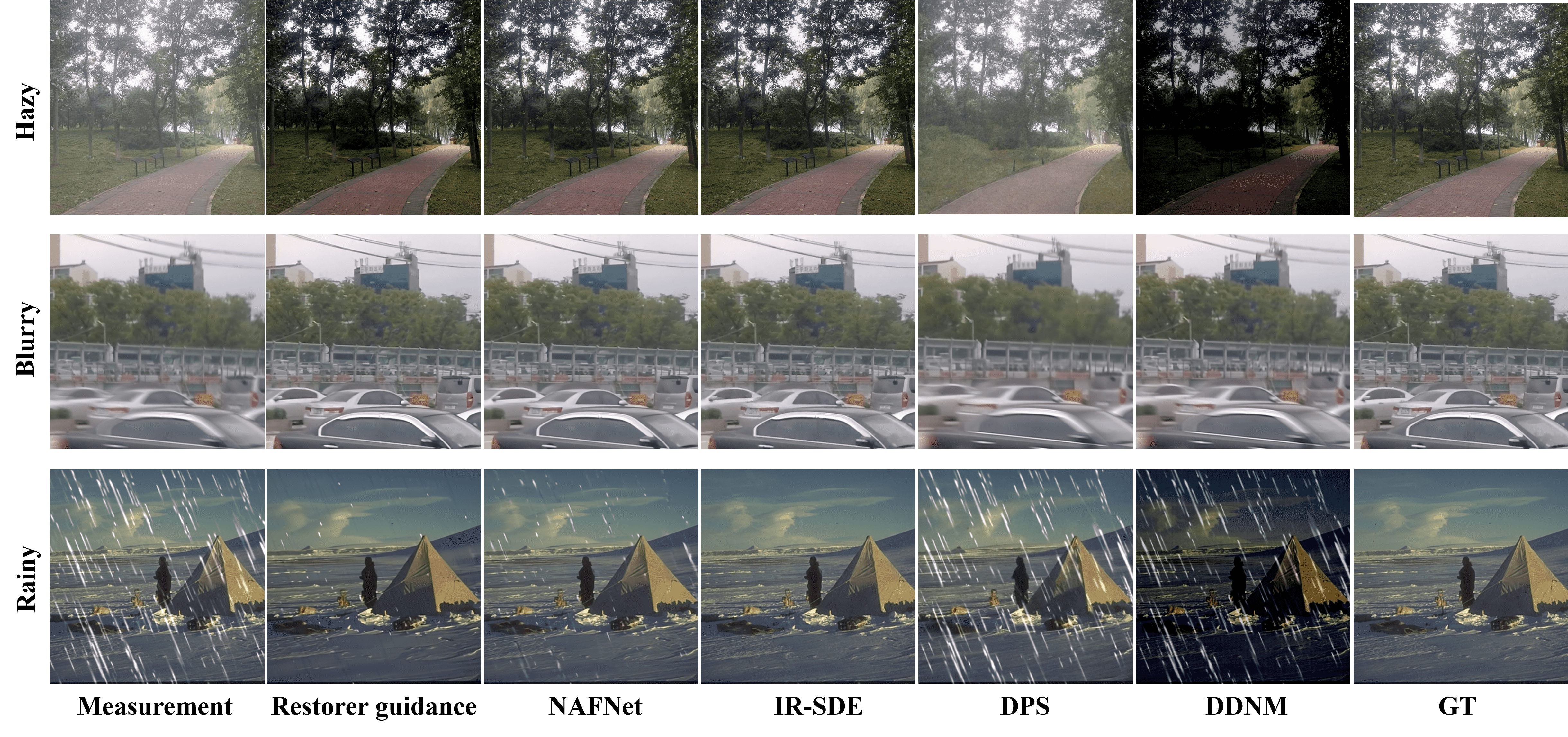}
    \vspace{-0.8cm}
    \caption{Visual comparison of restorer guidance with other inverse problem solvers on capricious unpredictable deterioration process, including image dehazing, rain streak removal, and motion deblurring. The restorer prototype is deployed with NAFNet for comparison. Best viewed zoomed in.}
    \vspace{-0.5cm}
    \label{fig:visual-app}  
\end{figure}

\begin{figure}[h]
    \centering
    \includegraphics[width=\textwidth]{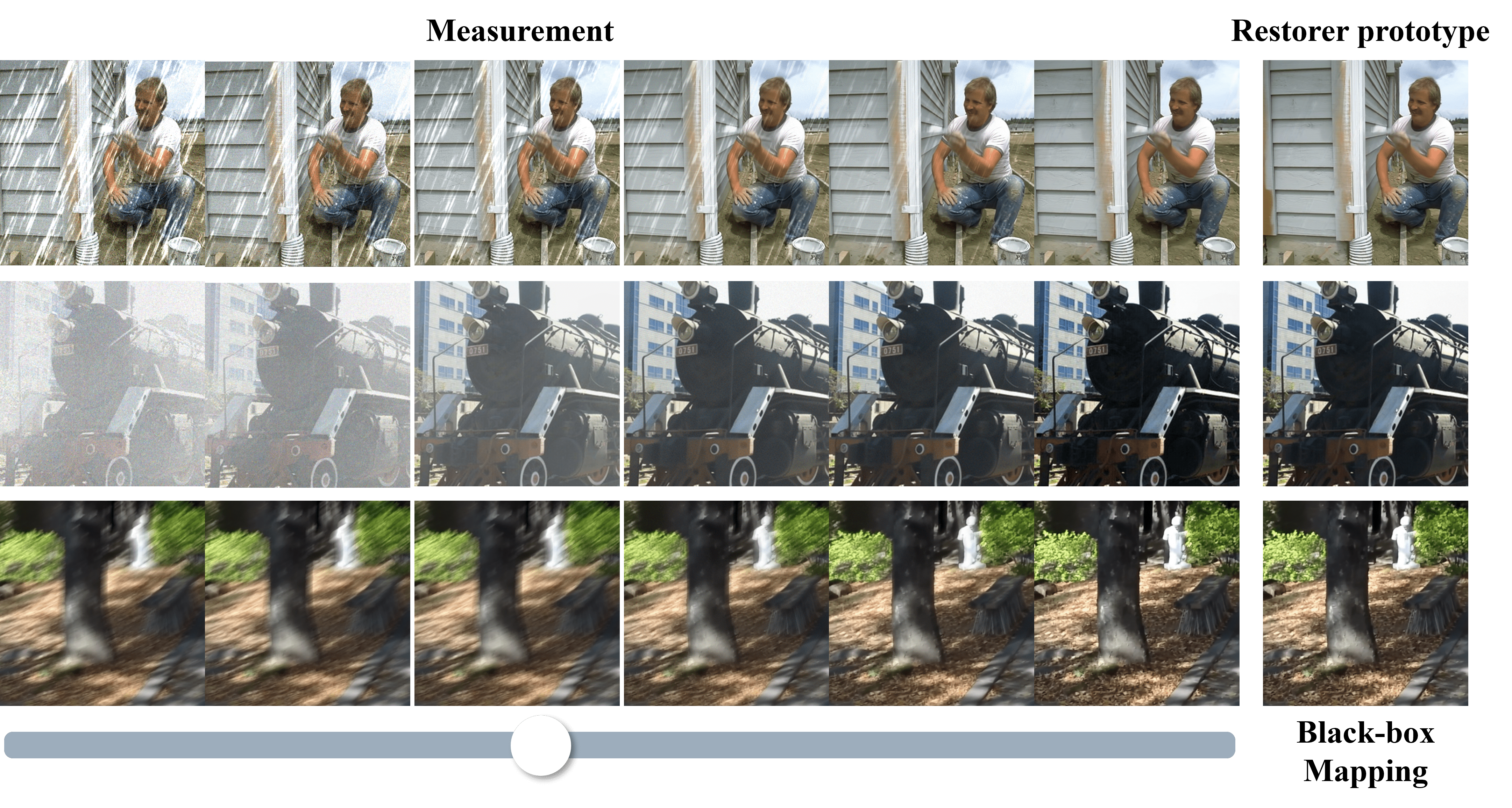}
    \vspace{-0.8cm}
    \caption{\textit{Restorer guidance} provides us a workbench to fabricate the restoration process more flexible and controllable with proficient deterioration expertise preserved in restorer rather than obstreperous black-box mapping.}
    \label{fig:ap3}  
\end{figure}

\begin{figure}[h]
    \centering
    \includegraphics[width=\textwidth]{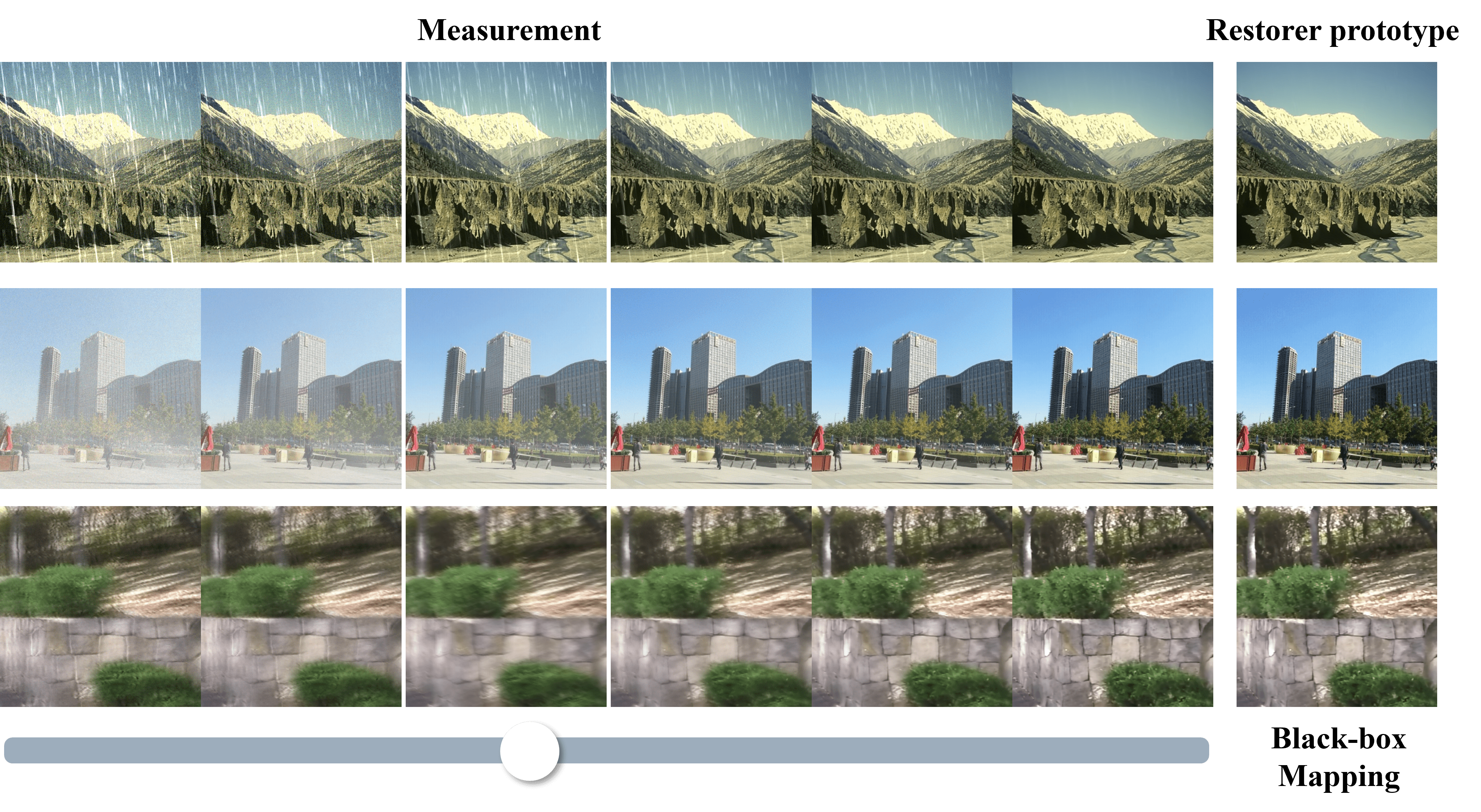}
    \vspace{-0.8cm}
    \caption{\textit{Restorer guidance} provides us a workbench to fabricate the restoration process more flexible and controllable with proficient deterioration expertise preserved in restorer rather than obstreperous black-box mapping.}
    \label{fig:ap4}  
\end{figure}

We provide additional visual results in Fig. \ref{fig:visual-app} to further illustrate the effectiveness and the behavior of the \textit{restorer guidance}. More visual results of the proficient deterioration control of the \textit{restorer guidance} are provided in Fig. \ref{fig:ap3} and \ref{fig:ap4}, where we can fabricate the restoration process more flexible and controllable with proficient deterioration expertise preserved in restorer.
It is supposed to accomplish the interconnected deterioration process with discretionary user inclination in the future.

\end{document}